\pdfoutput=1
\documentclass[11pt]{article}
\usepackage{authblk}
\usepackage[]{acl}

\usepackage{times}
\usepackage{latexsym}

\usepackage[T1]{fontenc}

\usepackage[utf8]{inputenc}

\usepackage{microtype}

\usepackage{graphicx}
\graphicspath{pics/}
\usepackage{caption}
\usepackage{boldline} 
\usepackage{multirow}
\usepackage{sidecap}
\usepackage{subcaption}
\usepackage{natbib}
\usepackage{amsmath}
\usepackage{amsfonts}
\usepackage{xcolor}
\usepackage{bm}
\usepackage{hyperref}
\usepackage{rotating}
\usepackage{adjustbox}
\usepackage{algorithm}
\usepackage{algpseudocode} 
\usepackage{array}
\definecolor{green}{rgb}{0.0, 0.5, 0.0}
\definecolor{red}{rgb}{0.82, 0.1, 0.26}
\definecolor{codegray}{gray}{0.9}

\newlength{\Oldarrayrulewidth}

\usepackage{booktabs}

\usepackage{listings}
\definecolor{codegreen}{rgb}{0,0.6,0}
\definecolor{codegray}{rgb}{0.5,0.5,0.5}
\definecolor{codepurple}{rgb}{0.58,0,0.82}
\definecolor{backcolour}{rgb}{0.95,0.95,0.92}
\lstdefinestyle{mystyle}{
  backgroundcolor=\color{backcolour}, commentstyle=\color{codegreen},
  keywordstyle=\color{magenta},
  numberstyle=\tiny\color{codegray},
  stringstyle=\color{codepurple},
  basicstyle=\ttfamily\footnotesize,
  breakatwhitespace=false,         
  breaklines=true,                 
  captionpos=b,                    
  keepspaces=true,                 
  numbers=left,                    
  numbersep=5pt,                  
  showspaces=false,                
  showstringspaces=false,
  showtabs=false,                  
  tabsize=2
}
\title{ Make Text Unlearnable: Exploiting Effective Patterns to Protect Personal Data }

\author{\textbf{Xinzhe Li}, \textbf{Ming Liu}, \textbf{Shang Gao}}
\affil{School of IT, Deakin University, Australia \\
\texttt{\{lixinzhe, m.liu,shang.gao\}@deakin.edu.au}}

\begin{document}
\maketitle

\begin{abstract}

This paper addresses the ethical concerns arising from the use of unauthorized public data in deep learning models and proposes a novel solution. Specifically, building on the work of \citet{huang2021unlearnable}, we extend their bi-level optimization approach to generate unlearnable text using a gradient-based search technique. However, although effective, this approach faces practical limitations, including the requirement of batches of instances and model architecture knowledge that is not readily accessible to ordinary users with limited access to their own data. Furthermore, even with semantic-preserving constraints, unlearnable noise can alter the text's semantics. 
To address these challenges, we extract simple patterns from unlearnable text produced by bi-level optimization and demonstrate that the data remains unlearnable for unknown models.  Additionally, these patterns are not instance- or dataset-specific, allowing users to readily apply them to text classification and question-answering tasks, even if only a small proportion of users implement them on their public content.
We also open-source codes to generate unlearnable text and assess unlearnable noise to benefit the public and future studies.
\end{abstract}

\section{Introduction}
With the increase in the prevalence of deep learning, public data has become more frequently used for developing predictive models. However, the use of unauthorized public data, such as tweets, raises ethical concerns. Furthermore, it is considered even more unethical to charge the public for services based on these models.
In addition to the ethical concerns, our research can help address privacy issues associated with the development of sensitive applications that impede public privacy. For instance, facial recognition systems can recognize individuals even when they are on the street \citep{hill2020secretive}. To prevent deep learning models from exploiting textual content and potentially predicting private information such as sentiments on sensitive topics \citep{kouloumpis2021-twitter,severyn2015-twitter}, political affiliations \citep{conover2011-twitter}, age, and gender of users \citep{farnadi-2018-use-profiling, suman-2021-user-profiling}, we propose making text unlearnable.
While \citet{huang2021unlearnable} proposed a process to make images unlearnable, our work extends this idea to generate unlearnable text using a gradient-based search approach.
% mohammad-2022-ethics: The malicious use of deep learning to identify demographics \citep{farnadi-2018-use-profiling, nicolas2020-profiling-overview, suman-2021-user-profiling} has led to unfair intervention for political voting or internet bullying. 

In our study, we investigate the performance of error-minimization modifications for text unlearning in three tasks: sentiment analysis, topic classification, and question answering. Sentiment analysis and topic classification can reveal users' interests, such as political leaning, while question answering can extract information from users' text. Due to data accessibility limitations and privacy concerns, we conduct our experiments on open data that is commonly used for academic purposes.

Our contributions include the adaptation of the bi-level optimization formulation from \citet{huang2021unlearnable} to text, and the development of a search procedure to modify text for (inner) error minimization. Our results show the efficacy of error-minimization modifications in making text unlearnable for all three tasks. However, the optimization process is impractical in real-world scenarios. Therefore, we extract two synthetic patterns from error-min modifications: label hints for text classification and an answer hint for question answering. These patterns can make text unlearnable and can be applied to any individual text without requiring a computationally expensive algorithm.

We also consider the effectiveness of these synthetic patterns in real-world scenarios. Our results show that they can be effective on models with different network architectures and training paradigms, including training from scratch and the pretrain-then-fine-tune paradigm. Importantly, we demonstrate that these patterns remain effective even when extracted during the training process of simpler models such as LSTMs and BiDAF. Moreover, they remain effective even when only a portion of users use them, and can be applied to one of the classes, which can be helpful in making one specific sensitive class unlearnable.

 % Our analysis shows that the effectiveness of error-min modifications comes from simple patterns, which can make models rely on them during training and cause little gradient update on model parameters.
\section{Background}
In this section, we will conduct an analysis of the existing privacy protection methods designed to safeguard against training deep learning models.
We will then proceed to explicate the bi-level optimization approach adopted in this study to generate unlearnable images. 
In the subsequent section, we will demonstrate the generalizability of this method to generate unlearnable text

\subsection{Privacy Protection}
The development of deep learning models with public data has raised concerns about privacy leakage. 
Several research directions have been proposed to address this concern.
Differentially-private techniques \citep{dwork2014algorithmic,chaudhuri-2009-dp,shokri-2015-dp, mcmahan-2018-dp, abadi-2016-dp} have been suggested as a solution to prevent the memorization of user-specific information during the training process. However, the application of such techniques requires users to trust those who collect their data. 
Another proposed approach is machine unlearning \citep{cao2015-unlearn}, which aims to remove the training impact of specific samples provided by users after the models have successfully learned from the data. 

Protection of textual messages against unauthorized neural natural language processing (NLP) models is critical. Especially, statistical features learned by these models can lead to the extraction of private informationextracted by hackers \citep{fredrikson2015model, carlini2020extracting} 
% lin-etal-2021-using
since DNNs can memorize private information such as name and address in training data. This paper concentrates on user-end solutions for privacy protection, exploring noise-addition approaches against unauthorized NLP models. While several noise-addition approaches have been proposed by the computer vision community against facial recognition models \citep{shan2020fawkes, cherepanova2021lowkey,huang2021unlearnable}, to the best of our knowledge, no similar work has been conducted in the NLP community.
% Rather than stop models from being developed, \citet{viejo-2012-user-profiling} prevents well-developed models from profiling users. Moreover, it only works on the application of user profiling, while our work explores a generic approach for different applications.
% \citep{} Protecting Open-Source Software from Unauthorized Neural Code Learning

\subsection{Formulating the Unlearnable Objective as a Bi-level Optimization Problem}
Consider a training set $\mathcal{D}={(x, y)}_{i=1}^{N}$, where the $i$-th instance consists of a text $x$ and its true label $y$ for classification.
A DNN $f(\theta)$, where $\theta$ is parameters of the model $f$, maps the input space $\mathbb{X}$ to the output pace $\mathbb{Y}$. The training objective is to minimize the loss function $\mathcal{L}$:
\begin{equation} \label{eq:train_objective}
\begin{split} 
\underset{\theta}{\arg \min } \mathcal{L}(f(\boldsymbol{x}), y)]
\end{split}
\end{equation}

\paragraph{Min-min Optimization by \citet{huang2021unlearnable}.}
\citet{huang2021unlearnable} nested the unlearnable objective within the training objective (Equation \ref{eq:train_objective}) to formulate a bi-level optimization problem:
\begin{equation} \label{eq:huang_unlearnable}
\begin{split} 
\underset{\theta}{\arg \min } \mathbf{E}_{(\boldsymbol{x}+\eta, y) \sim \mathcal{D}}[\underset{\eta}{\arg \min } \mathcal{L}(f(\boldsymbol{x}+\eta), y)], 
% \quad \text { s.t. } \quad {\eta}_{p} \leq \epsilon
\end{split}
\end{equation}
where a pixel-wise vector $\eta \in \mathcal{R}^{C \times H \times W}$ is optimized to minimize $\mathcal{L}$, , where $C, H, W$ are the numbers of channels, height and weight of images respectively.

They solved the outer objective with the common training routine, i.e., the gradient descent algorithm to iteratively optimize the model parameters $\theta$:
\begin{equation} \label{eq:train_nn}
\begin{split} 
\theta_{t+1} = \theta_{t} - \gamma \nabla_{\theta_t} \mathcal{L},
\end{split}
\end{equation}
where $\gamma$ is the learning rate.

For the inner objective, they nested another iterative process of projected gradient descent (PGD) \citep{madry2017towards} 
% basic iterative method (BIM) \citep{kurakin2016adversarial}
to optimize the noise $\eta$ (error-min noise) for each training sample (sample-wise noise) or each class (class-wise noise), which is a common routine to solve bi-level optimizations \citep{finn2017model,huang2020metapoison}. 
Equation \ref{eq:pgd} shows the one-step update:
\begin{equation} \label{eq:pgd}
\begin{split} 
\eta_{t+1} = \eta_{t} - \varepsilon \operatorname{sgn} \nabla_{\eta_t} \mathcal{L}(\Tilde{x}_t),
\end{split}
\end{equation}

where they obtained perturbed images via element-wise addition $\Tilde{x}=x+\eta$, and $\varepsilon \operatorname{sgn}$ performs a $\ell_{\infty}$ norm.

We detail the whole min-min optimization in Algorithm \ref{alg:alg2}.
\begin{algorithm}
	\caption{Generating Unlearnable Noise.} \label{alg:alg2}
	\begin{algorithmic}[1]
	 \Require {
	 neural network $f(\theta)$,
	 training set $\mathcal{D}$, 
	 test set $\mathcal{D}_\text{test}$,
	 training loss $L$, 
	 initialized noise $\eta$, 
	 num of training steps per modification $M$ }
	 \State num\_train\_steps $\gets 0$ ; test\_metric $\gets \text{null}$
	 \For {each batch $Z \in \mathcal{D}$}
 		\If {num\_train\_steps $\pmod M$ = 0}
 		    \State Evaluate the current checkpoint $f(\theta)$ on $\mathcal{D}_\text{test}$ to get new\_metric
	        \If {test\_metric=null $\lor$ new\_metric > test\_metric }
	            \State test\_metric = new\_metric
	        \Else 
	            \State \Return {the noise $\eta$}
	        \EndIf
	        \State \emph{Update noise $\eta$ via an error-min optimization} (images: Equation \ref{eq:pgd}) \label{alg2:milestone_generate}
		\EndIf
	    \State Apply current unlearnable noise for all $x \in Z$ (images: $\Tilde{x}=x+\eta$) \label{alg2:milestone_apply}
		\State $\theta \gets \theta - \gamma \nabla_{\theta} \mathcal{L}(Z)$
		\State num\_train\_steps $+= 1$
	 \EndFor
	\end{algorithmic}
\end{algorithm}

Unlike the original process, we add the exit condition when the evaluation metrics on test sets are unchanged for computational efficiency, which indicates the noise's effectiveness. \footnote{We would use accuracy for text classification tasks and F1 scores for question answering.} 
To generate unlearnable text, we replace the step \ref{alg2:milestone_generate} with a loss approximation search procedure, as demonstrated in the next section.
% , which would modify the steps \ref{alg2:milestone_apply} and \ref{alg2:milestone_generate}.

% \paragraph{Error maximization for adversarial attacks.} In contrast, Previous works \citep{goodfellow2014explaining, ebrahimi2017hotflip} use error maximization to modify text for adversarial attacks on well-trained models.

% The advent of pre-trained transformers (e.g., BERT \citep{devlin-etal-2019-bert}, RoBERTa \citep{liu2019roberta}) has revolutionized the NLP applications. Therefore, we would make pre-trained transformer models unlearnable during their common fine-tuning paradigm.

\section{Adaptation to Text}
In this section, we first formulate noise as discrete text modifications in contrast to pixel-wise vectors for images.
To adapt Algorithm \ref{alg:alg2} with text modifications, we use a search procedure (Algorithm \ref{alg:alg1}) to replace PGD optimization steps.

\subsection{Text Modifications}
Unlike images,  a textual input $x$ consists of a sequence of words ${w_1, w_2, ..., w_T }$, where $T$ is the number of words. A vocabulary $V$ consists of all the words.
Therefore, we define noise as substituting the word $w_p \in x$ indexed by the position $p$ with a word $s \in V$, denoting as $\eta=(p, s)$.

However, there are two problems:
1) The discrete operation $(p, s)$ is not differentiable:  
Since the noise $\eta$ for images is continuous, it is differentiable and can be optimized via gradient descent. However, we cannot use gradient descent to optimize $(p, s)$;
2) Modifying a single token may change the semantics of text (e.g., "I love you" to "I the you"), while a simple $\ell_{\infty}$ norm on noise for an image can make it imperceptible.

\subsection{A Search Procedure}
To solve the first problem, we approximate the loss change for all possible substitutions and search for a substitute word causing the lowest loss.
Specifically, each word $w$ can be transformed into a dense vector $e_w$ via a matrix $\mathbf{E} \in \mathcal{R}^{n \times m}$ parameterized in a DNN $f(\theta)$, where $n$ is the size of a vocabulary $V$ and $m$ is the size of each embedding vector.
% which is indexed by its position $p$ in $x$, 
We measure the loss change of substituting a word $w_p$ with another word $s \in V$ by the inner product of $e_s$ and the gradient of loss w.r.t. $e_w$ ($\nabla_{e_w} \mathcal{L}(x, y)$).
\begin{equation} \label{eq:unlearnable_approx}
\begin{split} 
    \underset{s}{\arg \min } \quad e_s^{\mathrm{T}} \nabla_{e_w} \mathcal{L}(x, y)
\end{split}
\end{equation}
The first-order approximation approach has been used for adversarial attacks \citep{wallace2019universal, wallace2020imitation,ebrahimi2017hotflip} with different implementations.

% We also apply a part-of-speech constraint to ensure the correct grammar.
For semantic preservation, we select the modified word $s$ from semantically similar words for each substitution. 
Following the setting of \citet{alzantot2018generating} for generating adversarial candidates, we calculate the cosine similarity between $w$ and $s$ and select candidate words within the threshold. 
We discuss the setting of the hyperparameters in Appendix \ref{app:hyperparameter}.

Besides, we only consider one modification $(p, s)$ for a text. For question answering, we exclude positions in answer spans.

\paragraph{Implementation.}
To search for a $(p, s)$ to minimize the training loss, we acquire the gradients for all the positions of the original example by one forward and backward pass, i.e., $\nabla_{x} \mathcal{L}=\nabla_{e_{w_1}} \mathcal{L}, ..., \nabla_{e_{w_T}} \mathcal{L}$.

Instead of searching over the vocabulary for each $w_p$, we efficiently approximate the loss changes for all the candidates $(P, S)$ by one matrix multiplication as Equation \ref{eq:approx_score}. We discuss the approximation errors in Appendix \ref{app:approximation_error}.
\begin{equation} \label{eq:approx_score}
\begin{split} 
    \mathbf{A} = \nabla_{x} \mathcal{L}^{\mathrm{T}} \mathbf{E},
\end{split}
\end{equation}
where $\nabla_{x} \mathcal{L} \in \mathcal{R}^{T \times m}$, and embedding matrix  $\mathbf{E} \in \mathcal{R}^{n \times m}$,

We then rank all the candidates according to the approximation scores $\mathbf{A} \in {T \times n}$ and select the one with the lowest score satisfying the constraints.
 
Algorithm \ref{alg:alg1} demonstrates the process of searching for a optimal $(p*, s*)$ for an instance $(x, y)$ at one iteration. 

\begin{algorithm}
	\caption{Error-min for Gradient-based Word Substitutions.} \label{alg:alg1}
	\begin{algorithmic}[1]
	    \Require{a neural network $f$ with $\mathbf{E}$,
	    training loss $\mathcal{L}$, and a sample $(x, y)$}
	    \State Generate $\nabla_{x} \mathcal{L}(f(x), y)$ 
		\State Generate approximation scores $A$ for all the candidate position/substitution pairs $(P, S)$ via first-order approximation 
		\State Sort $(P, S)$ in the ascending order of $\mathbf{A}$
		
		\For {each candidate modification $(p, s) \in (P, S)$}
		\If {$(p, s)$ satisfies all the constraints}
		\State \Return $(p, s)$
		\EndIf
		\EndFor
	\end{algorithmic} 
\end{algorithm}

\section{Experimental Settings}
This section will first introduce all our experiment's tasks, datasets, and models. 
We then demonstrate essential factors for generating unlearnable modifications. 

\subsection{Tasks and Datasets} 
\paragraph{Text classification.}
A neural network $f(\theta)$ takes a text $x$ and outputs a probability distribution over the output space $Pr(\hat{Y}| x)$ after normalizing by the Softmax function, i.e., $Pr(\hat{Y}| x) = \text{Softmax}(f(x))$.
$\mathcal{L}$ is defined as a negative log likelihood of $Pr(y| x, \theta)$ or a cross entropy between $Pr(\hat{Y}| x)$ and one-hot representation of the true label $y$. 

We choose two datasets to simulate real-world scenarios to identify users' sentiments and interests, each with training, validation, and test sets.
\begin{itemize}
    \item SST2: It contains movie reviews from the Stanford Sentiment Treebank (SST) dataset. Each sentence is labelled as either positive or negative sentiment. \citep{socher-etal-2013-recursive} 
    \item AG-News: This dataset divides news articles into four classes: world, sports, business, and sci/tech.
    It involves 10,800 training samples, 12,000 validation samples, and 7,600 test samples.
    It works as a proxy task to detect users' interests.
    % \item Twitter gender predictions: The dataset \footnote{https://www.kaggle.com/crowdflower/twitter-user-gender-classification} contains 11,194 samples from the Twitter's user descriptions. We follow \citet{gorman-bedrick-2019-need} to randomly split the data for training, validation and test by the ratio of 7:2:1. 
\end{itemize}

\paragraph{Question answering.} 
Given a passage of text $p$ and a question $q$, models aim to extract a correct answer span $a$ from $p$. Given $x=(p, q)$, $f(\theta)$ will output probability distributions for both the beginning and ending positions of the answer span $a$, denoting as $Pr_{\text{start}}$ and $Pr_{\text{end}}$.
The loss $\mathcal{L}$ is calculated by adding negative log likelihoods of $Pr_{\text{start}}$ and $Pr_{\text{end}}$.
We aim to prevent QA models from learning the passage when we maintain correct answers in the passage.

We use the Stanford Question Answering Dataset (SQuAD) v1.1 dataset \citep{Rajpurkar_2016}, which contains more than 100,000 question-answer pairs based on about 500 articles. 
%Crowd-workers generate the question-answer pairs. 
Since the SQuAD test set is unavailable, we use the validation/test splits from \citet{du-etal-2017-learning} derived from the original validation set.

\subsection{Models} 
To generate error-min modifications, we use LSTMs \citep{hochreiter1997long} ($\sim$ 3.8M parameters) for all the text classification tasks and Bidirectional Attention Flow (BiDAF) model \citep{sea2016-bidaf} ($\sim$ 2.5M parameters) for question answering. Specifically, BiDAF uses one bidirectional LSTM to represent each context and question respectively and applies an attention mechanism to generate two question-aware context representations with a dimension of $H$, where $H$ is the hidden size. 
A linear layer parameterized by a matrix $M^{H \times 2}$, followed by a softmax function, transforms them into the probability distributions $Pr_{\text{start}}$ and $Pr_{\text{end}}$ respectively.
% CNN \citep{kim-2014-convolutional}.
We use the 300-dimensional GloVe word vectors \citep{pennington-etal-2014-glove} for the above models.

To answer whether we can make text unlearnable when fine-tuning powerful pretrained language models, we evaluate BERT\textsubscript{BASE} with 110M parameters \citep{devlin-etal-2019-bert} for text classification
and RoBERTa\textsubscript{BASE} with 125M parameters \citep{liu2019roberta} for question answering.
In contrast to BiDAF, RoBERTa is pretrained to support a pair of sequences as inputs by concatenating them with a special token.

\subsection{Computational Considerations} 
Generating modifications by the min-min optimization is computationally expensive. Due to limited computational resources, we down-sample the training set for AG-News and SQuAD to validate the min-min optimization, i.e., using the first 3,200 articles and their categories of AG-News and 1,000 question-answer pairs from the SQuAD training set. 
However, we construct the vocabulary on the whole training data to avoid out-of-vocabulary when evaluating test data. 
Note that such size of SQuAD examples is not large enough to train a good QA model. However, we can evaluate the effectiveness of the min-min optimization by comparing model performance on clean and modified data.

Even so, we find that the algorithm \ref{alg:alg1} runs much slower on AG-News and SQuAD than SST2 since it is harder to find substitute words to satisfy the similarity constraint. We would not apply the constraint to AG-News and SQuAD. Since the text in these two datasets are much longer (19 for SST2, 43 for AG-News, and more than 100 for SQuAD), it is unlikely to change the semantics of a text by substituting one word. \footnote{Even so, running Algorithm \ref{alg:alg1} to generate one set of error-min modifications once costs around 4 hours for AG-News and more than 10 hours for SQuAD with RTX3080 (16GB).}

%maybe because these two datasets contain words with more diverse semantics and part-of-speech than those in SST2. 
% let alone the min-min required running the error-min algorithms multiple times on different model checkpoints.
% (2) control the trade-off between the convergence of model training and the error-min optimization for modifying training data. 

% Reviewer: very unclear here, need rewrite. so you are actually injecting noise to the original training data to train the model and return low performance on test data.
% ===========
% \citet{finn2017model}: $L(f_{\Tilde{x}})=L(f_{x-\alpha \dots \nabla_{\vect{x}} \mathcal{L}_{a d v}(\vect{x}, y) })$ 
% \paragraph{Why would the min-min optimization work?} 
% The inner minimization using the loss w.r.t. training data would decrease the training loss $\mathcal{L}$ by modifying clean data. Therefore, it would decrease the sensitivity of the outer minimization, since both optimizations have the same objective $\mathcal{L}$.
% ==============
% \paragraph{Problems.} Both minimizations are sloved by gradient-based methods, requiring the gradient of $\mathcal{L}$ w.r.t $\vect{\Tilde{x}}$ for the inner optimization and the gradient of $\mathcal{L}$ w.r.t $\theta$ for the outer optimization. The same problem as data poison is that the derivative of $\mathcal{L}$ w.r.t $\vect{\Tilde{x}}$ also relies on the training process for $f_{\theta}$

\begin{figure*}[h!]
        \centering
        \begin{subfigure}[b]{0.42\textwidth}
          \centering
          \includegraphics[width=0.95\textwidth]{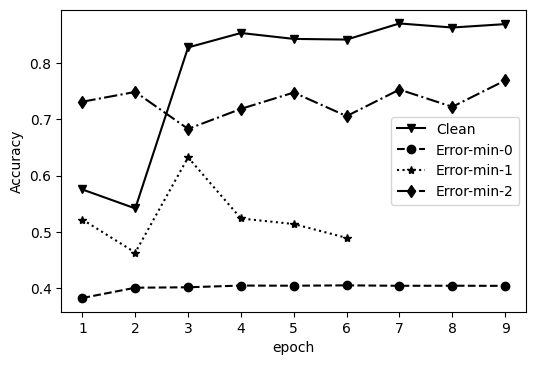}
            \caption[]%
            {{\small Test Accuracy on AG-News. }}  
            %  for Twitter gender prediction. We measure the test accuracy after each update of model parameters, and the loss reaches convergence within one epoch of training
            \label{fig:ag_news_acc}
        \end{subfigure}
        \begin{subfigure}[b]{0.42\textwidth}
            \centering
            \includegraphics[width=0.95\textwidth]{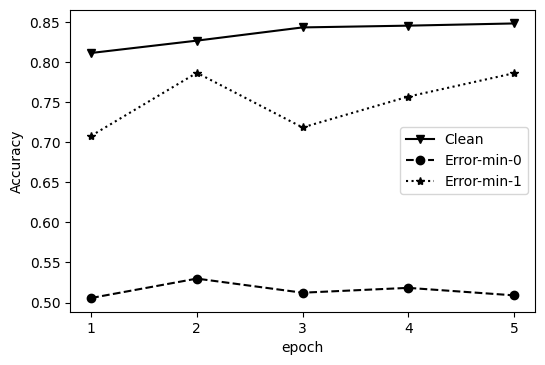}
            \caption[]%
            {{\small Test Accuracy on SST2. }}    
            \label{fig:sst2_acc}
        \end{subfigure}
        
        \vskip\baselineskip
        \begin{subfigure}[b]{0.42\textwidth}
            \centering
          \includegraphics[width=0.95\textwidth]{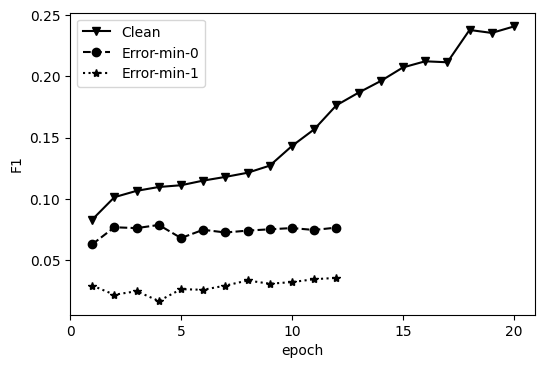}
            \caption[]%
            {{\small F1 scores on SQuAD. }}  
            %  for Twitter gender prediction. We measure the test accuracy after each update of model parameters, and the loss reaches convergence within one epoch of training
            \label{fig:squad_f1}
        \end{subfigure}
        \begin{subfigure}[b]{0.42\textwidth}
            \centering
            \includegraphics[width=0.95\textwidth]{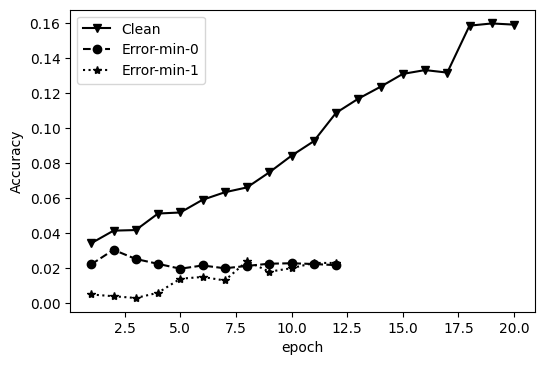}
            \caption[]%
            {{\small Exact Match on SQuAD. }}    
            \label{fig:squad_em}
        \end{subfigure}
        \caption[]
        {\small Test metrics under error-min modifications during the training.
        We train LSTM models for the classification tasks and BiDAF for SQuAD.
        Note that some lines halt in the middle due to early stopping.} 
        \label{fig:min_min_effectiveness}
\end{figure*}

\section{Effectiveness of Min-min Optimization}
In this section, we report the effectiveness of modifications generated via the min-min optimization and further analyze why min-min modifications are effective.

\subsection{Experimental Results}
The min-min optimization generates several sets of error-min modifications $(S_0, P_0), ..., (S_i, P_i), ..., (S_{N}, P_{N})$ at different training checkpoints (see step \ref{alg2:milestone_generate} in Algorithm \ref{alg:alg2}). For example, $\text{Error-min-i}=(S_i, P_i)$ is generated by Algorithm \ref{alg:alg1} after $M \times i$ training steps, which would be applied on the next $M$ training steps (see step \ref{alg2:milestone_apply} in Algorithm \ref{alg:alg2}) until $(S_{i+1}, P_{i+1})$ is generated.
$\text{Error-min-N}=(S_{N}, P_{N})$ is the final output from the min-min optimization.

We not only answer whether the final min-min modifications ($\text{Error-min-N}$) can make text unlearnable but also evaluate whether other sets of error-min modifications (e.g., $\text{Error-min-i}$) can be effective. 
Specifically, we apply each set of error-min modifications to the clean training data and optimize neural networks on the modified training data.
We then follow the strategy from \citet{huang2021unlearnable} to measure metrics on test samples during different training epochs. 
The min-min optimization over LSTM on SST2 generates three sets of error-min modifications (i.e., $N=3$), while two sets for SST2 and SQuAD. 

All the results in Figure \ref{fig:min_min_effectiveness} demonstrate that the $\text{Error-min-0}$ modifications effectively make text unlearnable. They are even more effective than the last error-min modifications for SST2 and AG-News.
With this, the bi-level optimization may be unnecessary to generate effective modifications, and one-step error minimization on randomly initialized DNNs can generate effective modifications.

\subsection{Analysis}
After exploring why $\text{Error-min-0}$ appears more effective in this section, we find that there exist simple, explicit patterns which correlate to the task-specific outputs (i.e., labels for text classification or answers for QA) to make text unlearnable.

Specifically, we first investigate whether substitute words in each set of error-min modifications correlate with labels. We divide all the substitute words for each class into bags of words (label-wise BOWs) and calculate the average Jaccard similarity between each pair of BOWs as Equation \ref{eq:jaccard_sim}. 
Table \ref{tab:jaccard_sim} shows that 
effective modifications (e.g., $\text{Error-min-0}$) present low similarity, which indicates that label-wise patterns may make text unlearnable.

% We then analyze the training dynamics to answer the question: why do simple patterns make text unlearnable?
% we then analyze the practicality of using such unlearnable modifications in real life

% It is distinct from class-wise noise for image classifier \citep{huang2021unlearnable}, which are explicitly defined for each class. 
\begin{equation} \label{eq:jaccard_sim}
\begin{split} 
    \text{Average Similarity} = \sum_{i=1}^{K} \sum_{j=i+1}^{K} \frac{|\text{BOW}_i \cap \text{BOW}_j|}{|\text{BOW}_i \cup \text{BOW}_j|}
    % \text{Score}(S, S_{\text{min-min}}) = \text{min}(\text{TermFreq}(w_i, S), \text{TermFreq}(w_i, S_{\text{min-min}})) \forall w_i \in S_{\text{min-min}}
    % denote $S$ as a bag-of-word models $S=\{(w_1,c_1), ..., (w_K, c_K)\}$ , where $(w, c)$ is a word/count pair.
\end{split}
\end{equation}
where $K$ is the number of classes/labels.
\begin{table}[ht!]
    \centering
    \begin{tabular}{llr}
    \toprule
     Task & Modifications & Value \\
    \midrule
        \multirow{3}{*}{AG-News}
        & Error-min-0 & 0\\ 
        & Error-min-1 & 0.08\\
        & Error-min-2 & 0.12\\
    \midrule
        \multirow{2}{*}{SST2}
        & Error-min-0 & 0\\ 
        & Error-min-1 & 0.36 \\
    \bottomrule
    \end{tabular}
    \caption{The average Jaccard similarity between each pair of bag of words by labels.}
    \label{tab:jaccard_sim}
\end{table}
We also find little sample-wise feature in each label-wise BOW. Specifically, we calculate the probabilities over all the substitute words. 
For example, $Pr_{BOW_0}(w)$ denotes the probability that the word $w$ appears amongst all the samples with the label indexed by $0$. 
We then rank the probabilities in descending order and cumulate the probabilities for the top 5 words. Figure \ref{tab:cumulative_probs} shows that we only need five words to make most of the examples unlearnable.
\begin{table}[ht!]
    \centering
    \begin{tabular}{llrrr}
    \toprule
     \multirow{2}{*}{Task} & \multirow{2}{*}{Labels} & \multicolumn{3}{c}{Error-min} \\
     & & 0 & 1 & 2 \\
    \midrule
        \multirow{4}{*}{AG-News} & World & 0.99 & 0.88 & 0.96 \\ 
         & Sports & 0.96 & 1 & 1 \\
         & Business & 0.91 & 1 & 0.92 \\
         & Science & 1    & 0.91 & 0.9 \\
    \midrule
        \multirow{2}{*}{SST2}
        & Negative & 0.6 & 0.73 & / \\ 
        & Positive & 0.87 & 0.69 & / \\
    \bottomrule
    \end{tabular}
    \caption{The cumulative probabilities of the top 5 substitute words.}
    \label{tab:cumulative_probs}
\end{table}

\begin{figure}
    \centering
    \includegraphics[width=0.45\textwidth]{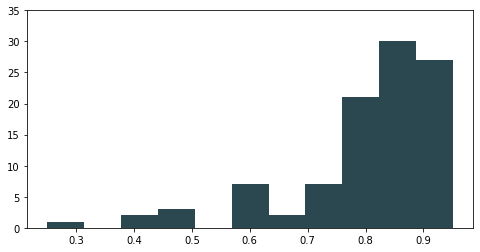}
    \caption{Distribution of relative positions to modify.}
    \label{fig:dist_P}
\end{figure}

We then investigate the distribution of positions $P$. We calculate the relative position $p_{rel}$ for each sample by dividing each position $p$ (the index of the modified word) by the length of the sentence $x$. Extremely,
$p_{rel}=0$ when modifying the first word, while $p_{rel}=1$ if the last word is modified.
Figure \ref{fig:dist_P} shows that text tends to be modified at the end. 
% In fact, all the examples are modified at the last two words. 
%In summary, we assume label-wise patterns inserted at the end can make text unlearnable in classification tasks.

We also find a simple pattern in the error-min modifications for SQuAD:
1) all the positions are identified within the one-word distance of the answers. 
2) Similar to text classification, the top 5 substitute words modify 98\% of 1000 samples. 

Therefore, we can reasonably hypothesize that the min-min optimization would generate noise with task-specific patterns to make text unlearnable, e.g., words correlating to labels for text classification or words to indicate the positions of answers for QA.

\begin{figure*}[h!]
        \centering
        \begin{subfigure}[b]{0.42\textwidth}
            \centering
          \includegraphics[width=0.95\textwidth]{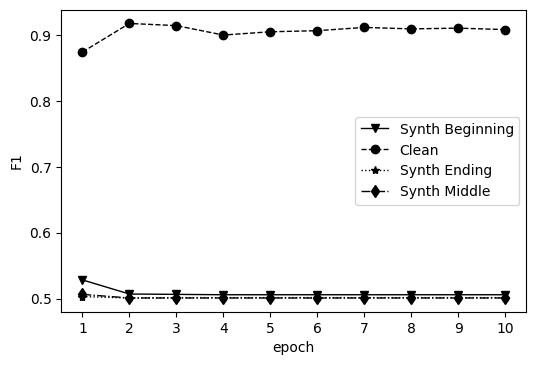}
            \caption[]%
            {{\small SST2. }}  
            \label{fig:bert_acc}
        \end{subfigure}
        \begin{subfigure}[b]{0.42\textwidth}
            \centering
            \includegraphics[width=0.95\textwidth]{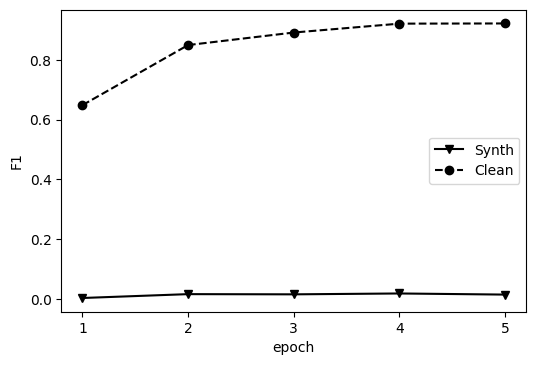}
            \caption[]%
            {{\small SQuAD }}    
            \label{fig:}
        \end{subfigure}
        \caption[]
        {\small The performance of synthetic features. We report test accuracy when fine-tuning BERT on SST2 and F1 scores when fine-tuning RoBERTa on SQuAD.} 
        \label{fig:sst2-bert}
\end{figure*}

% Unlearnable State: models learns the conditional probability distribution $Pr(Y|b)$, which is independent of $x$, i.e.,
% \begin{equation} \label{eq:unlearnable_state}
% \begin{split} 
% Pr(Y|b) = Pr(Y|b, x),
% \end{split}
% \end{equation}
% for each $(x, y) \in \mathcal{D}$.
% Since we cannot ensure the existence of such $b$, we use the KL divergence between $Pr(\hat{Y}|b)$ and $Pr(\hat{Y}|b,x)$ to evaluate the Equation \ref{eq:unlearnable_state}, i.e., the unlearnable state.
% The low KL divergence indicates that the trained models would strongerly correlate the biased feature $b$ with $y$.

\begin{table*}[t!]
    \centering
    \begin{tabular}{llp{8cm}}
        Dataset & Type & Examples  \\
         \hline
         \hline
         \multirow{4}{*}{SST-2} 
         & Negative label hint & This isn't a new idea[original: . modified:\textcolor{red}{\textit{\textbf{@}}}]
        \\
        \cmidrule{2-3}
        & Positive label hints & Part of the charm of Satin Rouge is that it avoids the obvious with humor and lightness[Original:. Modified: \textcolor{red}{\textit{\textbf{!}}}]
        \\
        \cmidrule{2-3}
        & Min-min & Part of the charm of Satin Rouge is that it avoids the obvious with humor and [Original:lightness Modified: \textcolor{red}{\textit{\textbf{commander-in-chief}}}].
        \\
        \bottomrule
    \end{tabular}
    \caption{Examples of Unlearnable Text}
     \label{tab:examples}
\end{table*}
\section{Manually Generating Simple Patterns} \label{sec:pattern}
In this section, we test the effectiveness of synthetic patterns according to the previous findings since it is difficult to use the min-min optimization in reality. First, it assumes that users can access model architectures and the whole training data (or at least a batch of instances). In real life, users can only access their portion of data and publish one instance (e.g., a tweet) once at a time. 
Besides, generating modifications with the min-min optimization is very computationally expensive. 

Hence we construct synthetic patterns, including class-wise symbols (\emph{label hints}) for text classification and a symbol surrounding the answer spans (\emph{answer hints}) for question answering.
Another benefit is that inserting such symbols maintains semantics without complicated constraints.

To show that the patterns can be generalized to other network architectures, we evaluate them by fine-tuning two popular pretrained transformers: BERT for text classification and RoBERTa for question answering.
Figure \ref{fig:sst2-bert} shows that these hints can effectively prevent DNNs from comprehending the text.
Surprisingly, class-wise symbols are effective at any position (the beginning/middle/end). Although we show experimental results with characters (e.g., "a", "b") as the hints, we can also achieve the same outcome by inserting an exclamation mark ("!") and an at sign ("@") at the end of positive and negative reviews respectively as label hints, which makes such patterns more imperceptible (See Appendix \ref{tab:examples} for examples). 

\paragraph{The patterns' effectiveness when only partial training instances can be modified.}
Since it may not be possible to let all users add the patterns,
we explore their effectiveness when applying such patterns to partial training data.

We randomly select a certain percent of training instances ($\mathcal{D}_{\text{partial}}$) and apply unlearnable patterns on them ($\mathcal{D}_{\text{unlearn}}$). 
To show the effectiveness of unlearnable patterns, we calculate the change in the test accuracy after adding $\mathcal{D}_{\text{unlearn}}$ into the training process. For comparison, we report the result by adding $\mathcal{D}_{\text{partial}}$.
As shown in Table~\ref{tab:percent}, models rarely learn useful information from $\mathcal{D}_{\text{unlearn}}$ compared to $\mathcal{D}_{\text{partial}}$.

\paragraph{Can we only make one class of examples unlearnable?}
We select one class in AG-News (i.e., the "World" category) and insert a symbol ("a") only on instances belonging to the "World" class.
A BERT model fine-tuned on such a dataset shows low accuracy on the test instances belonging to the "World" class (0.015) and high accuracy on others (0.93). Henceforth, users can make a sensitive class of data unlearnable by agreeing on a class-specific symbol.

% what is the minimum percentage of data required to make models achieve the unlearnable state? a small portion of users providing 80\% training data can be transformed into unlearnable text.

\begin{table}[!t]
    \centering
    \begin{tabular}{lrrrr}
        \toprule
            &\multicolumn{3}{c}{SST2} & SQuAD \\
            \hline
            & 95\% & 90\% & 80\% & 80\% \\
        \midrule
        $\mathcal{D}_{\text{unlearn}}$ & +1\% & +1\% & 0 & -9\% \\
        $\mathcal{D}_{\text{partial}}$ & +6\% & +4\% & +2\% & +12\%\\
         \bottomrule
        %  & 95\% & 90\% & 80\% & 0  \\
        %  Modify  
        %  & 0.86 & 0.88 & 0.89 & 0.91   \\
        %  Skip
        %  & 0.85 & 0.87 & 0.89 & 0.91 \\
    \end{tabular}
    \caption{The change of the test accuracy after adding $\mathcal{D}_{\text{unlearn}}$ or $\mathcal{D}_{\text{parital}}$ into the training process.
    We construct $\mathcal{D}_{\text{unlearn}}$ or $\mathcal{D}_{\text{parital}}$ with different percentages of training data.
    % We fine-tune BERT on SST2 for 10 epochs and SQuAD for 4 epochs to the convergence in all the cases.
    }
    \label{tab:percent}
\end{table}

% \paragraph{Another Case Study: Twitter Gender Prediction.}

\subsection{Why Do Simple Patterns Make Text Unlearnable?}
We consider simple patterns as biased features.
Without any biased feature, the gradient descent algorithm would optimize $\theta$ to approximate the conditional probability $Pr(y| x)$ by minimizing empirical errors of any training instance. 
When we embed a simple biased feature $b$ into $x$, the DNN would first learn $Pr(y| b)$. Many previous works \citep{he-2019-unlearn, branco-etal-2021-shortcutted} have found that deep learning tends to learn superficial patterns. As shown in our experiments, once the model learns such $Pr(y| b)$,  models have difficulty exploiting the semantics of the input $x$ during the latter training process since the performance on test data does not improve. This property coincides with shortcuts found in question answering \citet{lai-etal-2021-machine}.

\begin{figure}
     \centering
         \includegraphics[width=0.46\textwidth]{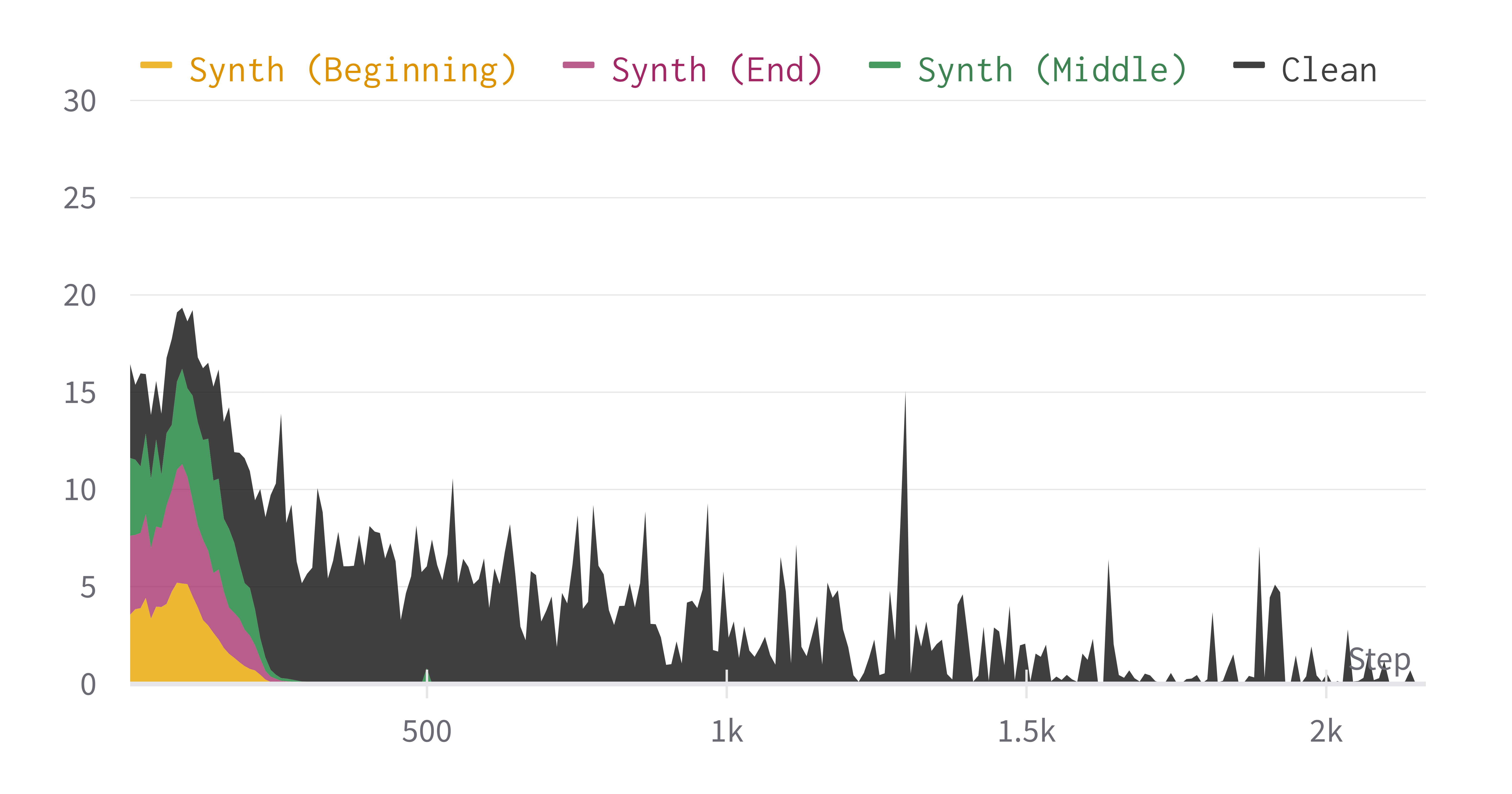}
        \caption{The change of gradient norms when we fine-tune BERT on SST2. Gradient norms shown in the stacked area chart.}
        \label{fig:bert_grad_norm}
\end{figure}

\paragraph{An unlearnable state.} 
We assume that there exists \textbf{\textit{an unlearnable state}} when models confidently correlate $b$ with model outputs, i.e., $Pr(y|b) \approx  1$, which would lead to $\mathcal{L} \approx  0$ for any input $x$ with $b$. 
Correspondingly, the forward pass would generate zero gradients to update the model during the backward pass. 
Since the model has no update according to the data, we can ensure that there is no information leakage.  
We verify this by tracing gradient norms during fine-tuning BERT on synthetic patterns.
Figure \ref{fig:bert_grad_norm} shows that the unlearnable state appears at about 250 iterations, where the model stops updating parameters. The same phenomenon occurs during training LSTM on error-min modifications (see Appendix \ref{app:grad_norm}).

% \begin{figure*}[h!]
%         \centering
%         \begin{subfigure}[b]{0.475\textwidth}
%             \centering
%           \includegraphics[width=0.95\textwidth]{pics/ag_news-lstm-grad-norm.png}
%             \caption[]%
%             {{\small AG-News. }}  
%             %  for Twitter gender prediction. We measure the test accuracy after each update of model parameters, and the loss reaches convergence within one epoch of training
%             \label{fig:ag_news_grad_norm}
%         \end{subfigure}
%         \begin{subfigure}[b]{0.475\textwidth}
%             \centering
%             \includegraphics[width=0.95\textwidth]{pics/sst2-lstm-grad-norm.png}
%             \caption[]%
%             {{\small SST2. }}    
%             \label{fig:lstm_grad_norm}
%         \end{subfigure}
        
%         \caption[]
%         {\small Tracing gradient norms during training.
%         We train LSTM models on SST2 and AG-News. Note that the area for Error-min-0 modifications (in Green) is too small to be visible.} 
%         \label{fig:grad_norm}
% \end{figure*}

\section{Conclusion}
By adapting min-min optimization, we develop an approach to expose vulnerabilities of deep learning to make text unlearnable.
To overcome the limitation of requiring knowledge of models and training data, we extract simple patterns (e.g., label hints and answer hints) from the min-min optimization to make text unlearnable. 
Although our experiment explores patterns for text classification and question-answering tasks, the pipeline potentially works for any NLP task.

% There may be other sensitive, linguistic forms for unlearnable objective: syntactic structure, commonsense, text style. Second, exploring unlearnable text on text generation models. This is also closely related to fact check in tasks like text summarization and machine translation.

\paragraph{Reproducibility.}
To ensure the effectiveness of unlearnable modifications, we slightly tuned the training hyperparameters to achieve well-trained models, such as setting maximum gradient norms and early stopping according to validation sets. 
We open-source codes with configuration files, which contain hyperparameters regarding model architectures (e.g., the number of layers), batching (e.g., data sampling), and training setups (e,g., learning rate).
Since these files are configurable in JSON format, future works can easily reproduce and extend the experiments.

\section{Limitations} \label{sec:limit}
% Experimental data in this paper may deviate from real-life text from users. For example, the SQuAD dataset consists of formally written Wikipedia articles, while the real-life text from users are written in a casual style. 
The main concern is that debiased techniques may remove simple biased features. However, to our knowledge, most debiased techniques \citep{rathore2021verb} can only remove biases across a concept subspace (e.g., the bias direction for gender) in the embedding space. Another setup of data debiasing, e.g., \citet{he-2019-unlearn}, requires hypothesized biases to train biased models and is limited to tasks with known hypothesized biases (e.g., lexical overlap for NLI). Also, they remove biased examples rather than identify biased symbols (e.g., label hints). However, we still expect future works to consider other complicated patterns beyond symbol insertions or word substitution.

% \paragraph{Future works.} 
% Therefore, in future work, we would like to explore the effect of debiasing techniques. 
% If simple features in our paper cannot work well under debiasing techniques, we may consider other types of biased features to construct unlearnable text, such as shortcuts \citep{branco-etal-2021-shortcutted,lai-etal-2021-machine,bai2021attention}. They contain superficial features, and some of them are very invisible and normally exist in the real-world data.

% For example, naive unlearnable features can be identified by a bias classifier \citep{he-2019-unlearn}, which predicts $y$ using only the biased feature $Pr(Y|b(x))$. 

% The goal is to prevent models from making predictions relying on the spurious feature $b(x)$ which may not exist during test time, i.e., $p(y|b(x)) \neq q(y|b(x))$ where $p$ and $q$ represent probability distributions for the training data and test data respectively. Our goal is to encourage so as to unlearn generalized information $g(x)$, especially private information (e.g., gender, age).
 
% For question answering, shortcut symbols near the answer would let models only learn to answer the questions according to the shortcut symbols rather than sophisticated reasoning and syntactic or semantic understanding of text. However, it may have risks of letting humans easily identify the important information if they know the shortcut symbols.
% Therefore, we may need define more invisible patterns as shortcuts.

\bibliography{anthology,custom}
\bibliographystyle{acl_natbib}

% \clearpage
\appendix

\section{The Change of Gradient Norms} \label{app:grad_norm}
\begin{figure}[ht]
         \centering
         \includegraphics[width=0.5\textwidth]{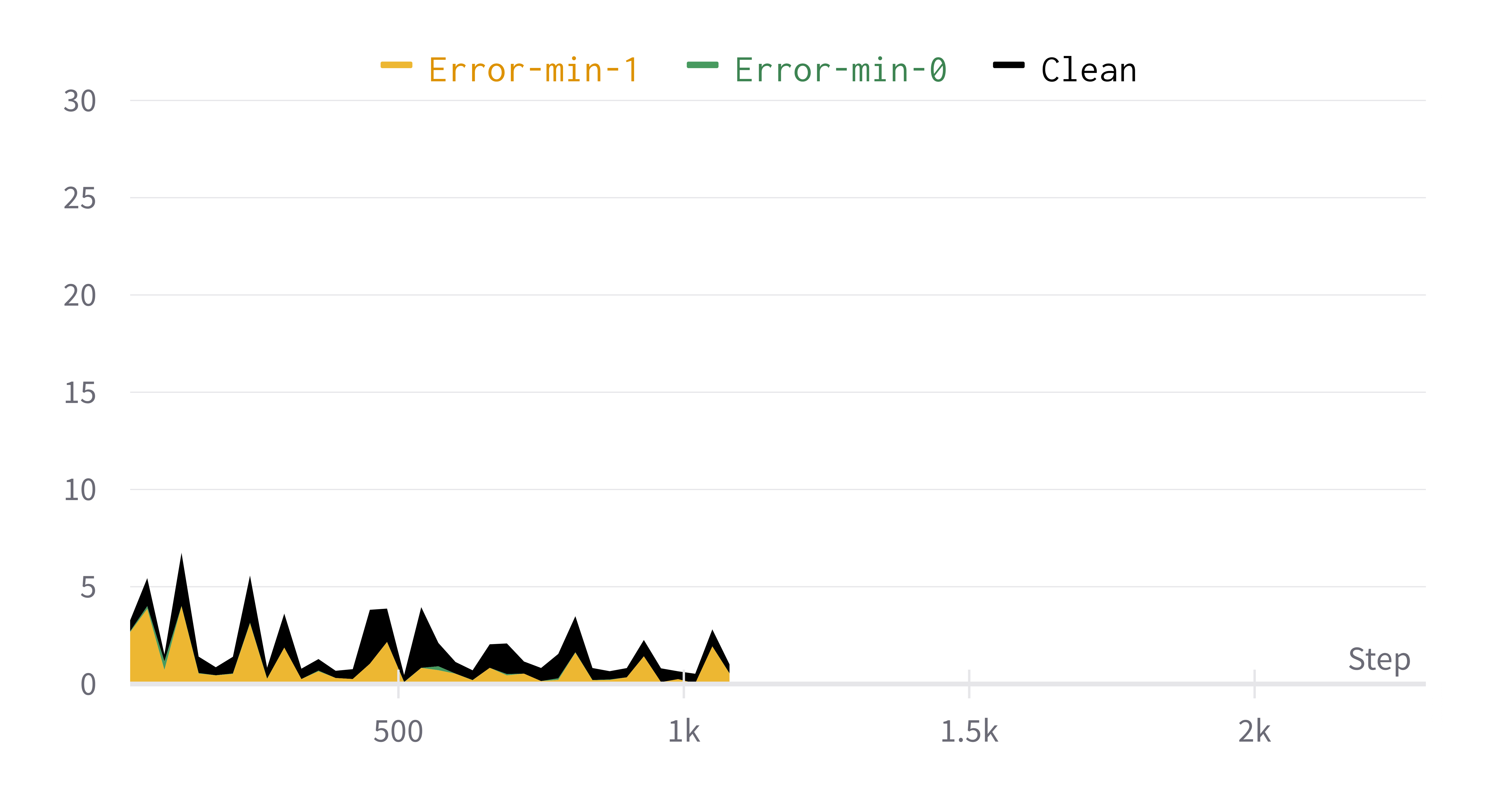}
        \caption{Training LSTM on SST2 from scratch. Note that the area for $\text{Error-min-0}$ modifications (in Green) is too small to be visible. Gradient norms shown in the stacked area chart.}
        \label{fig:lstm_grad_norm}
     \end{figure}

Figure \ref{fig:lstm_grad_norm} shows gradient norms with error-min modifications and further proves the argument. The set of the $\text{Error-min-0}$ modifications with label-wise patterns (see Table \ref{tab:jaccard_sim}) has almost zero gradients during training. It even has a small gradient update in the first few steps. It may be because the randomly initialized models can easily learn class-wise patterns, while BERT has to overcome its pretrained priors.

\section{Hyperparameter Setting} \label{app:hyperparameter}
\paragraph{The interval of optimizing the error-min noise $M$.} 
If $M$ is too small, the test accuracy after another $M$ iterations easily plateaus due to insufficient model update, which causes the early stop of the min-min process. 
% Henceforth, the unlearnable modifications may not be reliable.
On the other hand, a large interval will linearly increase the computational complexity. Specifically, since we use modifications for batches of instances in the next $M$ training iterations, error-min optimization needs to be run for $M \times B$ instances, where $B$ is the batch size.
 % (Step \ref{alg2:milestone1} in Algorithm \ref{alg:alg2})
 
Hence, we set $M=30$ for text classification tasks and a smaller $M$ ($10$) for SQuAD because of a larger batch size and longer sequence lengths to train SQuAD models.

% dynamic M: update unlearnable text with a certain improvement of evaluation metrics/loss 
% no train: update on randomly parameterized models
% Algorithm \ref{alg:alg2} shows how the interval $M$ works during the min-min optimization.

\paragraph{The threshold of cosine similarity.}
We set the threshold to 0.5 to follow the work \citep{alzantot2018generating} for generating adversarial noise.
 The effect of the threshold: 
Increasing the threshold can help find more semantically similar words (even synonyms), as specified in \citet{Mrksic2016CounterfittingWV}. For example, when we use this threshold, the word "award-winning" is identified to replace "charming". However, by increasing the threshold to 0.9, the substitute word becomes "lovely". 
However, Algorithm \ref{alg:alg2} runs much slower by denying most of the high-ranked candidates and leads to noise that is hard to make data unlearnable.
Also, it stops us from deriving general unlearnable patterns via qualitative analysis of substitute words. For example, the cumulative probabilities in Table \ref{tab:cumulative_probs} would be smaller due to more varying substitution sets. 

\section{ Errors of Approximating Loss Changes } \label{app:approximation_error}
Generally, in our experiment, Equation \ref{eq:approx_score} can always approximate the loss change in a correct direction, in our case, leading to the decrease of the actual loss. 
Specifically, the errors of the approximate loss change depend on the state of the models (the outcome of the outer minimization). For example, the results (the loss on the original SST2 training instances/the loss on the modified instances/the approximate loss change) for a randomly initialized LSTM would be 0.6931/0.6833/-0.0004, while, at the other extreme, the results for the LSTM checkpoint which has converged on our label hint are 0.4457/0.0782/-0.0012 or 0.4905/0.0714/-0.0379.

% \section{EMNLP2022 Reviews}
% The last nine pages show the reviews from EMNLP2022.

% \paragraph{What parts of the paper have been revised?}
% According to the reviews, we made the following revisions to the paper.
% \begin{itemize}
%     \item discussing the approximation errors in Appendix \ref{app:approximation_error};
    
%     \item discussing hyperparameter setting in Appendix \ref{app:hyperparameter};
    
%     \item emphasizing the practicality of simple patterns: We 1) modify the abstract and introduction, 2) extend the conclusion and 3) use more invisible patterns as label hints and answer hints and show the result in \S \ref{sec:pattern};

%     \item adding examples for various types of unlearnable text in Appendix \ref{tab:examples};

%     \item discussing how debiasing techniques can potentially remove unlearnable noise in \S \ref{sec:limit}.
    
% \end{itemize}

% \onecolumn
% \includegraphics{review1.png}
% \includegraphics{review11.png}
% \includegraphics{review2.png}
% \includegraphics{review21.png}
% \includegraphics{review3.png}
% \includegraphics{review31.png}
% \includegraphics{review4.png}
% \includegraphics{review41.png}
% \includegraphics{review42.png}

\end{document}